\documentclass{article}
\usepackage{cite} % Make references as [1-4], not [1,2,3,4]
\usepackage{url}  % Formatting web addresses  
\usepackage{ifthen}  % Conditional 
\usepackage{multicol}   %Columns
\usepackage[utf8]{inputenc} %unicode support
\usepackage[left=0.75in,right=0.75in]{geometry}

\usepackage{amsfonts}
\usepackage{subfig}
\usepackage{amsmath}
\usepackage{latexsym}
\usepackage{setspace}
\usepackage{graphicx}
\usepackage[tworuled,linesnumbered,noend,noline]{algorithm2e}
\usepackage{rotating}
\usepackage{multirow} 
\usepackage{multicol}
\usepackage[english]{babel}
\usepackage{blindtext}
\usepackage{array}% http
\title{FGPGA: An Efficient Genetic Approach for Producing Feasible Graph Partitions}

\author{ Md. Lisul Islam$^1$, Novia Nurain$^1$, Swakkhar Shatabda$^1$ and M Sohel Rahman$^{2,3,\dagger}$\thanks{$^{\dagger}$On a Sabbatical
leave from BUET.}
\\$^{1}$Department of Computer Science and Engineering, United International University 
\\$^2$Department of Computer Science and Engineering, Bangladesh University of Engineering and Technology
\\$^3$Department of Informatics, King's College London\\
Email: $\{lisul,novia,swakkhar\}$@cse.uiu.ac.bd, %swakkhar@cse.uiu.ac.bd, 
msrahman@cse.buet.ac.bd
}

\begin{document}

\maketitle
\begin{abstract}
Graph partitioning, a well studied problem of parallel computing has many applications in diversified fields such as distributed computing, social network analysis, data mining and many other domains. In this paper, we introduce FGPGA, an efficient genetic approach for producing feasible graph partitions. Our method takes into account the heterogeneity and capacity constraints of the partitions to ensure balanced partitioning. Such approach has various applications in mobile cloud computing that include feasible deployment of software applications on the more resourceful infrastructure in the cloud instead of mobile hand set. Our proposed approach is light weight and hence suitable for use in cloud architecture. We ensure feasibility of the partitions generated by not allowing over-sized partitions to be generated during the initialization and search. Our proposed method tested on standard benchmark datasets significantly outperforms the state-of-the-art methods in terms of quality of partitions and feasibility of the solutions. 
\end{abstract}

%\vspace{-0.1cm}

\section{Introduction}
Graph partitioning, a well-known problem of computer science and engineering, is widely used and studied in diversified practical and theoretical applications. Applications include parallel/distributed computing (load balancing of computations), scientific computing (fill-reducing matrix re-orderings), EDA algorithms for VLSI CAD (placement), data mining (clustering), social network analysis (community discovery), pathological and biological network analysis (detection of cliques), pattern recognition, and relationship network analysis. The goal of graph partitioning problem is to divide the vertices of a graph into sets of specified sizes, so that few edges cross between sets (i.e., minimizes a cut metric). 

\begin{figure}[!t]
\centering
\includegraphics[width=9cm,height=70mm]{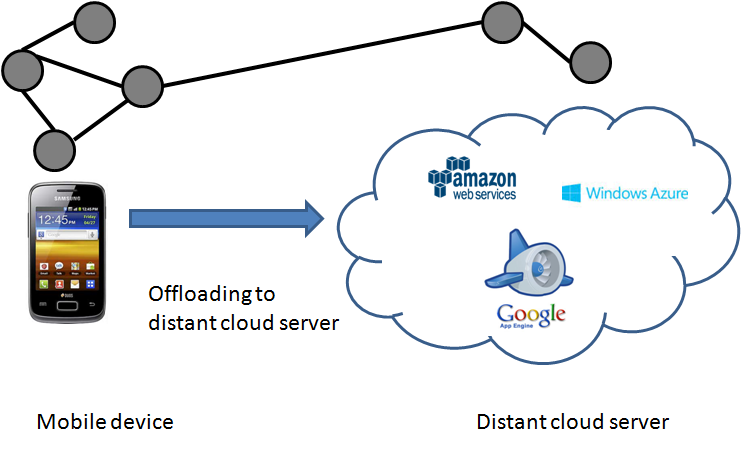}
\caption{Offloading of parts of software components from mobile device to distant cloud server}
\label{fig:1}
\vspace{-0.7cm}
\end{figure}

Recently, the graph partition problem has gained importance due to its gleaming applications in mobile cloud computing such as offloading parts of a software from the mobile hand set to a more resourceful server (Fig. \ref{fig:1}). Such offloading poses challenges to both cloud user and cloud vendor while allocating different software components to machines in the cloud minimizing the required bandwidth. A special case of such deployment optimization occurs in {\it cloudlet} \cite{1,2}, where multiple mobile devices access nearby static resourceful computers linked to a distant cloud through high speed wired connections. A global optimization is required taking into account all the software components of all devices. Moreover, a fast algorithm would also be required to adopt with the changes of optimal software deployment in the cloud with respect to time.

\ifx
In this paper, we focus on determining an efficient graph partition approach to partition the different components of a software application to deploy on a number of interconnected machines in the cloud with different capacities such that the partition minimizes the cost of communication between the components. \fi 

Such problem can be modeled as graph partitioning problem. Here, the weighted graph of software components has to be partitioned into a number of partitions, and each partition will be deployed in the available machines of the cloud. In this paper, we focus on determining an efficient graph partition approach to partition the different components of a software application to deploy on a number of machines with different capacities in the cloud. Here, our goal is to determine partitions such that each partition minimizes the cost of communication between the components. This problem can be viewed as an ILP (Integer Linear Programming) problem. ILP solver (IBM ILOC CPLEX \cite{3}) could be used to determine an optimal partition of the graph. However, the time and resource requirement depend on the cardinality of the graph. Both computation time and resource utilization grows exponentially with the increase in the graph size. Therefore, heuristic approaches are needed to find a good solution faster. Many heuristics of different nature (spectral \cite{4}, combinatorial \cite{5}, evolutionist \cite{6,7}, etc.) have been developed to provide an approximate result in a reasonable (and, one hopes, linear) computational time. Others existing approaches of graph partitioning such as Kernighan-Lin algorithm \cite{9} fails to scale to large scale graph data. Beside, all the existing methods partition the graph in to a predefined number of parts of equal sizes. In the context of mobile cloud computing, the number of partitions (i.e., the number of machines on which the components of the software will be deployed) is not predefined. Moreover, we need to take into account the diverse capacity of the machines in the cloud while deploying the software components. Therefore, existing graph partitioning algorithms cannot be directly applied to resolve such problem.

The authors of \cite{8} have introduced a graph partitioning algorithm based on simulated annealing \cite{14} for software deployment in the cloud. However, they consider only the homogeneous machines. Moreover, their proposed solution tends to give infeasible solution in some cases. Therefore, in this paper, we introduce an efficient genetic approach for producing feasible graph partitions. We name our approach as FGPGA. To resemble real scenarios of mobile cloud computing, our proposed method takes into account the existence of heterogeneity of machines along with different capacities of machines in the cloud. Moreover, feasible partitions of the components of software are ensured by discarding over sized partitions generated during the search. To summarize, in this paper, we make the following contributions: 
\begin{itemize}
\item We introduce an efficient genetic approach for producing feasible graph partitions named FGPGA to partition
different components of a software application on to a number interconnected machines in the cloud minimizing the communication cost
\item We consider heterogeneity and capacity constraint of the machines to resemble real scenarios
\item We guarantee feasible partition by disallowing over-sized partitions during the search, and
\item Finally, we perform extensive experiments to prove the efficacy and efficiency of our proposed approach.
\end{itemize}

Section~\ref{secProblem} formally describes our problem and necessary background required. Section~\ref{secRelW} reviews related works. In Section~\ref{secMethod}, we describe our proposed FGPGA method. Section~\ref{secResults} provides a detailed experimental evaluation of our algorithm compared with some existing simulated annealing algorithm and tests the efficacy and efficiency of our method. And finally in Section~\ref{secCon}, we draw a conclusion with future work.

\section{Preliminaries\label{secProblem}}
In this section, we provide a formal definition of the graph partitioning problem addressed in this paper and a brief description of genetic algorithms.
\subsection{Problem Model}
We adopted the problem model described in \cite{8}. However, the authors in \cite{8} used only homogeneous machines and assumed the machine graph to be complete and all edges in the machine graph to have same weight. In our problem, we are given two graphs: application graph $G_A$ and the machine graph $G_M$. The application graph, $G_A=(V,E)$ is an undirected weighted graph where the vertices of the graph represents distributed software units of deployment in the system and the edges corresponds to the communication cost between pair of software units. The vertices of this graph are also weighted. Each vertex $v_i \in V$ is associated with an weight $r_i$. Weight of a vertex corresponds to the amount of resource required for this unit deployment. Examples of such resource is CPU power. The adjacency matrix $W$ for this graph $G_A$ contains the weights for the edges, For each $w_{ij} \in W$, $w_{ij}$ represents communication overhead between software component $v_i$ and software component $v_j$. 

The infrastructure of the distributed system is also modeled as an weighted undirected graph $G_M=(M,L)$, where $M$ is the set of heterogeneous machines and $L$ represents the communication links between each machines. Each machine, $m_i \in M$ has a maximum capacity $C_i$ that represents the capacity of machine $m_i$ in terms of resources (i.e. CPU power). Communication link costs are found from the adjacency matrix $B$ for this graph $G_M$. For a link $l_{ij}$ between machine $m_i$ and machine $m_j$ corresponds to a communication cost, $b_{ij} \in B$. 
 
Now, the problem is to find a mapping $\phi:V\rightarrow M$ of the vertices in $V$ of $G_A$ to the vertices in $M$ of $G_M$. The mapping $\phi$ assigns each of the software component $v \in V$ to a machine $m \in M$. The objective of this mapping or partition is to minimize the graph cut size (GCS) defined as following:
\begin{equation}
GCS=\underset{i,j}{\Sigma} w_{ij}\times b_{\phi(v_i)\phi(v_j)}\times h_{ij}
\label{equGC}
\end{equation}  

Here, $h_{ij}$ defined depending whether two deployment units $i$ and $j$ are assigned to same machine or different. The definition is as follows:

\begin{equation}
h_{ij}=\begin{cases}
0, \text{ if } \phi(v_i)=\phi(v_j)\\
1, \text{ if } \phi(v_i)\neq \phi(v_j)
\end{cases}
\end{equation} 
 
Since the machines to which the units are deployed have a maximum capacity, we need another constraint that defines the feasibility of the partitions. This is a hard constraint defined in the following equation:  
\begin{equation}
\forall k: \underset{i}{\Sigma} r_i \times \Phi(i,k) \le C_k
\label{equFGP}
\end{equation}

Here $\Phi(i,m)$ is a binary valued function that returns true only if software unit, $i$ is assigned to machine $m$. Formally,
$$\Phi(i,k)=\begin{cases}
0, \text{ if }, \phi_(v_i)\neq m_k\\
1, \text{ if }, \phi_(v_i)=m_k
\end{cases}$$ 

In this model, Equation~\ref{equGC} defines the objective function for a balanced partition and the hard constraint defined in Equation~\ref{equFGP} ensures feasibility of the partitions.
\subsection{Genetic Algorithms}
Genetic Algorithms are population-based search methods that resemble the natural phenomena of biological evolution. Genetic Algorithms are widely used for different search optimization problems in different fraternity. Genetic algorithms maintains a set of solutions known as \emph{population}. Generally, it starts with a pool of initial random solutions which is called the initial population. Each individual in the population are encoded by a set of properties which are called chromosome or genotype which can be altered for attaining diversification in the population. It then follows an iterative process in which each of these iteration is called a \emph{generation}. In each generation, the population are then allowed to evolve using different operators which also mimics the natural process of biological evolution like \emph{mutation}, recombination or \emph{crossover} and survival of the fittest. In each generation, the fitness of each individual is evaluated. Generally the fitness is the value of the optimization function being considered to be solved. The more fit individuals are usually selected to breed and generate new fitter individuals in the population. Thus a new generation of population is `breeded' and are used in the next generation. This process of evolution are continues until a sufficient number of generations has been produced or a satisfactory level of fitness value has been attained.
\section{Related work}
\label{secRelW}
%\vspace{-0.2cm}
Graph partitioning has variuos applications in diversified research areas, such as parallel computing, circuit layout and the design of many serial algorithms. The term graph partitioning refers to partition the vertices of a graph into a certain number of disjoint sets of approximately the same size, so that a cut metric is minimized. Graph partitioning is a NP-complete \cite{11} problem and that there is no approximation algorithm with a constant ratio factor for general graphs \cite{11}. Because of this theoretical limitation, numerous heuristic algorithms for graph partitioning have been developed during the past decades that generate high-quality partitions in very little time.

Motivated by the problem of partitioning electronic circuits onto boards a heuristic method is produced in \cite{9}. This method is known as the Kernighan-Lin (KL) algorithm. The Kernighan-Lin algorithm subsequently improved in terms of running time by Fiduccia and Mattheyses \cite{12}. The spectral bisection method is another popular method, which is based on the spectrum of the graphs Laplacian matrix \cite{7,13}. However, this method has chances to stuck in local optima. In order to escape from the local optima these methods can be combined with different stochastic methods such as simulated annealing \cite{14}, particle swarm optimization \cite{15}, or ant colony optimization \cite{16}. Besides, all these methods do not scale to large scale graph data.

One recent approach that has greatly accelerated the partitioning of large graphs is the use of multilevel techniques \cite{10}. In multilevel approach, the graph is approximated by a sequence of increasingly smaller graphs. This graph is then partitioned using a spectral method \cite{17} and this partition is propagated back through the hierarchy of graphs. A variant of the Kernighan-Lin algorithm is applied periodically to refine the partition. Well-known software packages based on this approach include Jostle \cite{18}, Metis \cite{19}, and Scotch \cite{20}.

Recently graph partitioning algorithms are combined with different known techniques to develop new heuristics. Chardaire et al. use a PROBE (Population Reinforced Optimization Based Exploration) heuristic \cite{21}. Here, the authors combine greedy algorithms and genetic algorithms with KL refinement. Besides, Loureiro and Amaral introduce a greedy graph growing heuristic deploying a local refinement algorithm \cite{22}.

All these above graph partition methods consider a fixed and predefined size of partitions while partitioning the graphs. This characteristic restricts the usages of traditional methods directly for cloud computing. Hence, the authors of \cite{8}, proposed a graph partitioning algorithm based on simulated annealing for mobile cloud computing. Here, the authors consider the different capacities of machines along with the variation in the number of the machines on which the components of software are to be deployed. However, their approach provides infeasible solution for some cases. 

In this paper, we propose an efficient genetic algorithm for producing feasible graph partitions considering both the number and capacities of heterogeneous machines in the cloud. Extensive experimental result proves the efficacy and efficiency of our proposed method.

\section{FGPGA: Our Proposed Method\label{secMethod}}
Our proposed FGPGA algorithm is formally presented in Algorithm~\ref{algoMain}. The population of the first generation in our algorithm contains only feasible individuals which are initialized randomly. Our algorithm will terminate at convergence and convergence is achieved when, for a given span of generation, there has been noticed no substantial amount of improvement on the quality of the global best solutions. For each generation, our algorithm selects individuals using tournament selection to take part in the recombination(also known as cross-over) to produce offspring to be included in the next generation. A probabilistic mutation is also performed on each of the newly breed individual. Individual with best fitness value is always retained to the next generation to ensure elitism. To maintain diversification among the individuals in the search space and to recover from stagnation along the process of evolution, we initiate twin removal and random restart procedure periodically. Rest of this section are devoted to describe various components of FGPGA. Table \ref{table1} summarizes all the necessary parameters along with their values.
\begin{table}[ht]
\caption{Parameters used}
\centering
\begin{tabular}{c c } % centered columns (4 columns)
\hline\cline{1-2} %inserts double horizontal lines
Parameter & Value \\ [0.5ex] % inserts table
\hline % inserts single horizontal line
Population Size & 20 \\
Number of Generations run & 3000-6000  \\
Similarity Threshold & 0.95 \\
Random Restart Interval & 50 \\
Twin Removal & 100 \\
Tournament Size & 5 \\
\hline %inserts single line
\end{tabular}
\label{table1} % is used to refer this table in the text
\end{table}
\subsection{Encoding}\label {sub:encod}
In our method, we have encoded every individual in a population by $V$ number of genes. Here, $V$ is the number of component in a software application. Each of these genes in an individual is initialized randomly from an uniform distribution of the range $[1,M]$ where $M$ is the total number of heterogeneous machines in machine graph. So, an individual $X$ can be represented as the ordered list of $V$ number of genes, where each gene $X_i$ represents an assignment of a software component into a machine. 
\begin{equation*}
	X = \left\{ X_1,X_2,X_3,\ldots,X_V \right\} 
\end{equation*}
Here, $X_i$ represents the machine to which $component_i$ has been assigned.
\subsection{Initialization}\label{sub:ini}
Each individual in the population represents an assignment of component of the application software into a physical heterogeneous machine which has got some capacity constraints. So, if a machine is assigned to host a component or set of components whose aggregated capacity requirement is more than the capacity of the respective machine, then we can call such set of assignment invalid or declare that individual as an non-feasible solution of the graph partitioning problem. So, during the initialization of the population, we looked for solutions or individuals in the feasible search space only, thus creating a population consisting feasible individuals only. For each gene in the genotype of an individual, we repeatedly looked for a machine with enough capacity to host the $i^{th}$ component of the application software. A sketch of the pseudo-code for initialization procedure is given below:

\begin{algorithm}
%\SetAlgoLined
\DontPrintSemicolon
%\KwData{Individual X}
$M$ = total number of machines in the infrastructure\;
\For{each gene $X(i)$ in the genotype of $X$}
{
	find a machine $m$ randomly chosen from $U(1,M)$ such that $free(m)\geq W(i)$ \;
	set $X(i)$ to $m$ \; 
}
\Return $X$
\caption{Initialization (Individual $X$)\label{algoInit}}
\end{algorithm}

\subsection{Greedy Mutation}
\begin{figure}[h]
\begin{center}
\includegraphics[width=0.35\textwidth]{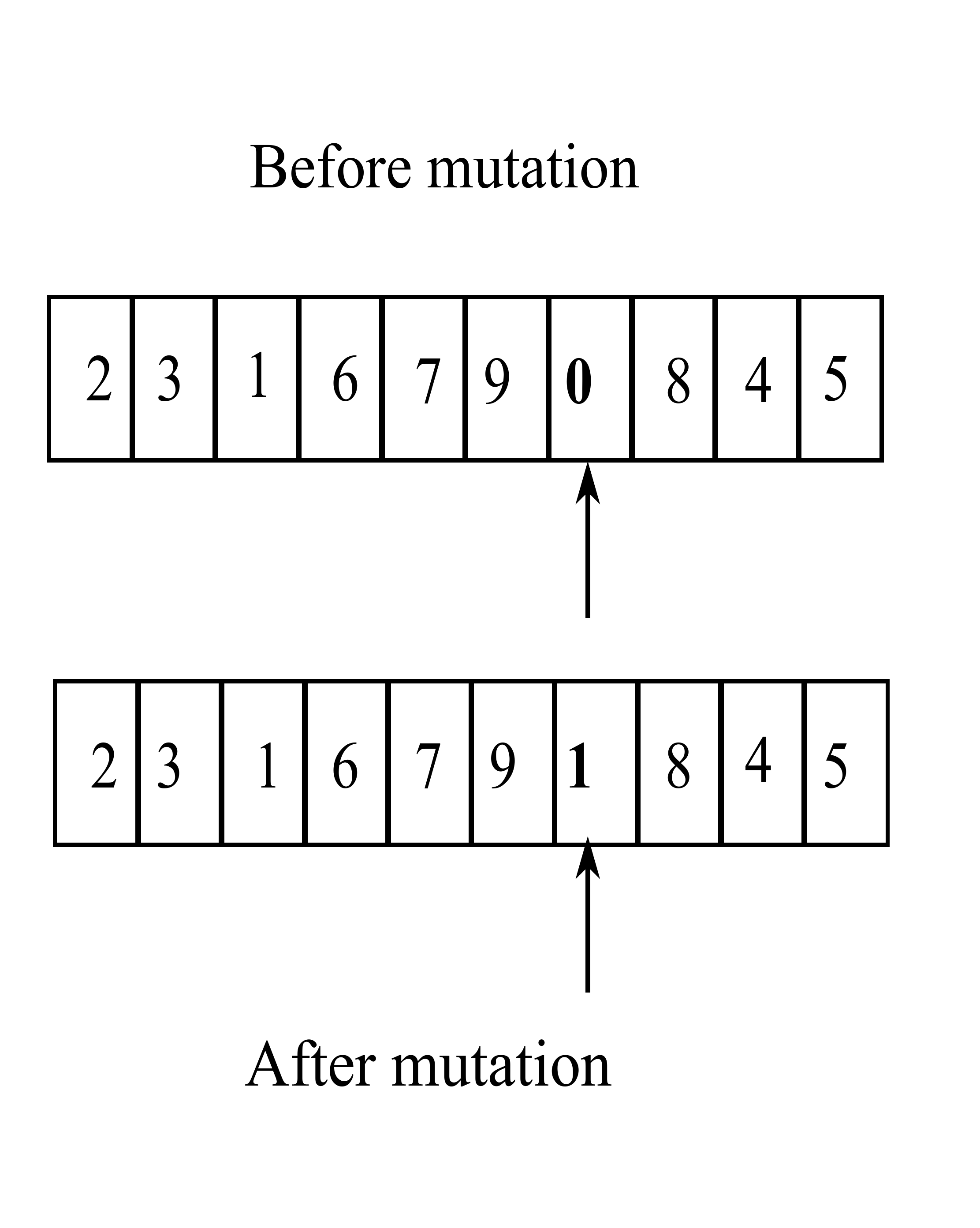}
\end{center}
\caption{An example of generic mutation process\label{fig:mut}}
\end{figure}
In our proposed algorithm, we adopted and introduced a greedy mutation strategy where we choose the new value of a particular gene of an individual optimally. We randomly alter a particular gene by trying out $r$ different values for that gene and finally retain the value from those $r$ different values that makes the individual fittest while still keeping it in the feasible region of search space. An example of greedy mutation is demonstrated  in Figure \ref{fig:mut}. Here, $r$ is set to the number of machines in the infrastructure. As greedy mutation is computationally cumbersome, we make selection between the random mutation and greedy mutation with a probability, $GreedyMutationRate(=0.8)$. If no feasible individual is found by mutating the currently selected gene, we then try out a different gene in the chromosome to mutate. The algorithm for greedy mutation is given in Algorithm~\ref{algoGM}.
\begin{algorithm}
\DontPrintSemicolon
%\KwData{Individual $X$}
set $r$ = number of machines \;
\For{each gene $X(i)$ in the genotype of $X$}
{
	S = set of $r$ random values for gene $X(i)$ from the range $[1,r]$\;
	find $v \in S$ for which fitness(X) is minimum and $X(i=v)$ will be a feasible solution\;
	set, $X(i) = v$\;
}
\Return $X$
\caption{Greedy Mutation (individual $X$) \label{algoGM}}
%\end{spacing}
\end{algorithm}
\subsection{Cross-over}
\begin{figure}[h]
\begin{center}
\includegraphics[width=0.5\textwidth]{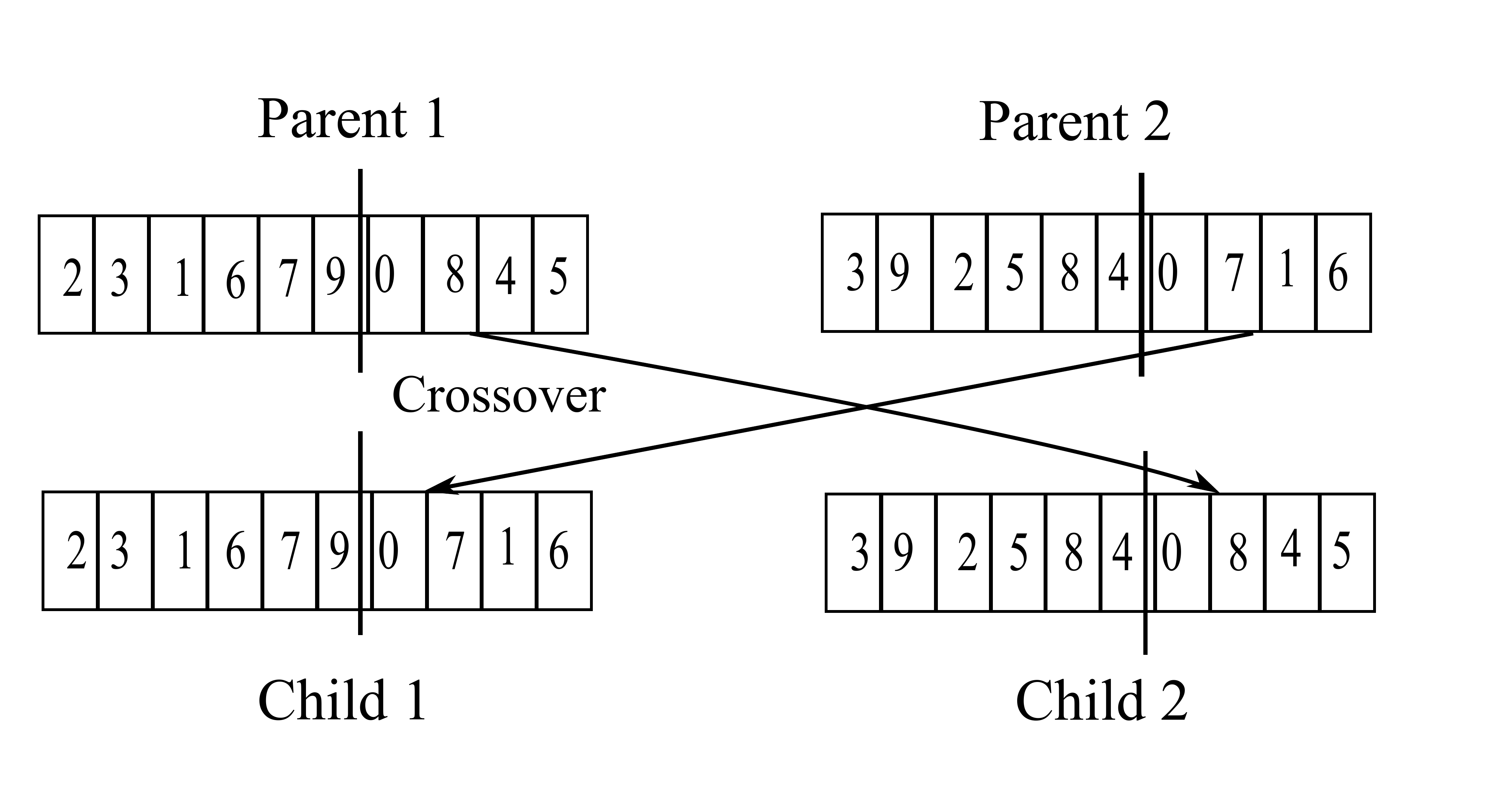}
\end{center}
\caption{An example of crossover process\label{fig:cross}}
\end{figure}
Cross-over operator aids the genetic evolution to exploit the fitter solutions found in the previous generations and generates new individuals by allowing fitter individuals to breed among themselves. To select individuals that will act as parents and take part in the recombination to produce offspring, we adopted tournament selection with the tournament size being equal to 5. Then, genes of the two parents are recombined using One-Point Crossover to generate a new set of gene for each of the newly formed offspring. Figure \ref{fig:cross} depicts an example of cross-over process. We tried out different cross-over point until we end up feasible offspring. If no feasible off-springs are found using the current parents, we then try out different parents by using tournament selection to generate feasible offspring. The algorithm is outlined in Algorithm~\ref{algoCO}.
\begin{algorithm}

\DontPrintSemicolon

%\KwData{Individual $X_1$, Individual $X_2$}

%\KwResult{Individual $X_{new}$}

$X_{new1}$ and $X_{new2}$ be two new offspring\;

\For{i=1 to number of genes in the chromosome of $X_1$}
{
	randomly choose a crossover point $k$\;
	generate $X_{new1}$ and $X_{new2}$ considering $k$ as crossover point\;
	\If{atleast one of $X_{new1}$ or $X_{new2}$ is feasible}
	{
		\Return $X_{new1},X_{new2}$\;
	}
}
\Return null
\caption{Cross-Over (individual $X_1$, individual $X_2$)\label{algoCO}}
\end{algorithm}

\subsection{Fitness Evaluation} \label{sub:ev}
We have ensured feasibility in the initialization, mutation and crossover. So, the search process will always explore within the feasible search space region. We have evaluated the fitness of an individual $X$   
by using equation ~\ref{equGC}.

\subsection{Twin Removal}
We have removed and reinitialized individuals with identical genetic information. We have defined the similarity measure between two individuals as the hamming distance of their corresponding genotype. Similarity between individual $X_1$ and $X_2$, Similarity($X_1$,$X_2$) is defined below as:
\[
  \frac{\text{number of $i$ such that $X_{1}(i)$ not equal to $X_{2}(i)$ }}{\text{size of the genotype}}
\]
We declare $X_1$ and $X_2$ as twins, if they have more than $95\%$ similarity in their genotype and  reinitialize randomly one of them to a feasible solution. We have run the this Twin Removal procedure after every 100 generations of evolution.
\begin{algorithm}
%\begin{spacing}{0.8}
\DontPrintSemicolon
%\SetAlgoLined
$similarityThreshold = 0.95$\;
\For{each pair of individuals $(X_i,X_j)$ in the population}
{
	\If{$Similarity(X_i,X_j) \geq similarityThreshold$}
	{
		declare $X_i$ and $X_j$ as Twins\;
		reinitialize $X_j$\;
	}
}
\caption{Twin Removal}
%\end{spacing}
\end{algorithm}

\subsection{Random Restart}
If the algorithm does not improve the fitness of the best individual within a significant number of generations, we reinitialize $50\%$ individuals of the population with random values within the feasible search space. We have evaluated the improvement over the immediate past 50 generations. If the improvement in the last 50 generations is less than or equal to a threshold, $t = 0.001$, we initiate the random restart procedure.

\begin{algorithm}
\DontPrintSemicolon
$nonDiverseSteps=0$\;
$nonImprovingSteps=0$\;
$GreedyMutationRate = 0.8$\;
Intialize the population, $P$ randomly\;
\While{Termination criterion are met}
{
	%find the individual $X_{best}$ with best fitness %in the $pop_i$\;
	$P_{new}=\{globalBest\}$\;
	\For{each Individual $X \in P$}
	{
		$\langle X_1,X_2\rangle = \mathsf{tournamentSelection}(P$)\;
		%$X_2 = \mathsf{tournamentSelection}(P$)\;
		$X_{new} = \mathsf{crossOver}(X_1,X_2$)\;
		\eIf{$Fitness(X_{new1})	\geq Fitness(X_{new2})X$}
		{
			add $X_{new1}$ to $P_{new}$\;	
		}
		{
			add $X_{new2}$ to $P_{new}$\;
		}
	}
	\For{each Individual $X$ in $P_{new}$}
	{
		\eIf{$rand(0,1) \leq GreedyMutationRate$}
		{
			$\mathsf{greedyMutate}(X)$\;
		}
		{
			$\mathsf{randomMutate}(X)$\;
		}
	}
	find the individual $X_{best}\in P_{new}$ with best fitness \;
	\eIf{$\mathsf{fitness}(globalBest) < \mathsf{fitness}(X_{best}) $}
	{
			{
			$globalBest=X_{best}$\;
			$nonImprovingSteps=0$}\;
	}
	{
		$nonImprovingSteps++$\;
	}
	%Replace any Individual $X$ with $X_{best}$ in the $pop_i$\;
	\eIf{ $nonDiverseSteps \ge twinRemovalInterval$ }
	{
		activate $\mathsf{twinRemoval}(P_{new})$ procedure\;
		$nonDiverseSteps=0$
	}
	{
		$nonDiverseSteps++$\;
	}
	\If{$nonImprovingSteps \ge randomRestartInterval $}
	{
		activate $\mathsf{randomRestart}(P_{new})$ procedure\;
		$nonImprovingSteps=0$
	}
	$P=P_{new}$\;
}
	\Return $globalBest$\;
\caption{FGPGA()}
 \label{algoMain}
\end{algorithm}

\section{Experimental Results\label{secResults}}
We have implemented FGPGA in Java programming language using JDK 1.6 and have run our experiments on an Intel(R) Core(TM)2 Quad CPU @ 2.40 GHz with 4GB RAM running Windows 7 operating system.

\subsection{Dataset Generation}
To evaluate the performance of our algorithm we generate test graphs following method similar to described in \cite{8}. However, the authors in \cite{8} assumed the infrastructure to be homogeneous. As a result, they considered $\forall i,j  b_{i,j}=1$, which means the communication between the machines in the cloud infrastructure are all uniform and the machine graph is a complete graph. Such instances are not realistic since they do not consider the heterogeneity of the infrastructure. We modify the graph generation technique to generate test cases that generates graphs with different node sizes and machine graph with heterogeneous communication links between them. 

We generated sparse graphs with the Eppstein
power law generator. The weights of the vertices and edges are taken from exponential distributions with the parameter, $\lambda$ equal to 0.1 and 0.005 respectively. For the machine graph, first we determined the capacity needed for the deployment and multiply it by 1.5 to ensure feasible infrastructure. Then machines were generated with capacity chosen randomly from $\{100,200,\cdots,800\}$. Machine graph is also generated as a sparse graph with edge weights generated using an exponential distribution with parameter, $\lambda=0.005$.
\subsection{Algorithm Comparison}
We compared our method with the simulated annealing method proposed in \cite{8}. In \cite{8}, a number of algorithms was proposed and applied to solve the graph partitioning problem. Among all the algorithms simulated annealing algorithm produced higher quality solutions for most of the graph instances that they used. We implemented the simulated annealing procedure described in their paper and compared it with FGPGA. In their simulated annealing \cite{8}, they allow infeasible moves or assignments fo vertices to partitions that are already over-sized and allow the capacity constraint to be violated. In their paper, they \cite{8} mention that due to the decrease in temperature in later epochs the rate of taking infeasible moves will decrease also and thus produce valid solution when the algorithm terminates. However, there is no guarantee for producing feasible solutions, once an infeasible move is taken. We observed that, our implementation of simulated annealing algorithm failed to produce any feasible solutions for most of the graph instances. It allows an infeasible solution at a higher temperature and further moves are not able to alter that solution to revert back to feasibility. To tackle this problem, we slightly modified the simulated annealing algorithm by not allowing the moves that turns any particular partition or machine to be over-loaded or the capacity constraint to be violated. Rather, a random move is allowed with the same metropolis condition instead of the infeasible move. Detail of the rest of the algorithm can be found in \cite{8}. Rest of the parameters were kept similar as described in the original paper. Since simulated annealing take long to converge, we allowed it to finish the epochs as described in \cite{8}. However, FGPGA were run for 6000 generations for smaller instances ($\le 500$ nodes) and for 3000 generations for largers instances ($\ge$ 600 nodes).    
\subsection{Results}

We report best and average graph cut size for the best partitions found by two algorithms simulated annealing and FGPGA in Table~\ref{table2}. Each algorithms were run 10 times for each of the graph instances and the average is reported in the table. Each row corresponds to the results for a particular graph instance. Values in bold faced fonts indicates the better quality partitions with lower graph cut size. Table~\ref{table2} clearly shows that FGPGA is able to produce better quality partitions in all the cases.  
\begin{table*}[ht]
\caption{Best and average graph cut size for different sized graph instances produces by two algorithms.}
\centering
\renewcommand*{\arraystretch}{1.4}
\begin{tabular}{c| c |cc |cc} % centered columns (4 columns)
\hline\cline{1-6} %inserts double horizontal lines
Graph & Number of vertices & \multicolumn{2}{c|}{FGPGA} & \multicolumn{2}{c}{Simulated Annealing}  \\ 
\cline{3-6} % inserts single horizontal line
Instance&$|V|$&{best}&{avg}&{best}&{avg}\\
\hline
1&100&{\bf 0}&{\bf 0}&{\bf 0}&{836134.6925}\\
2&200&{\bf 0}&{\bf 0}&{828585.8149}&{15307879.98}\\
3&300&{\bf 15652.12951}&{\bf 1155697.937}&{342046.9027}&{26553315.85}\\
4&400&{\bf 1841375.678}&{\bf 4483987.405}&{2.41E+07}&{7.79E+07}\\
5&500&{\bf 0}&{\bf 0}&{4.95E+05}&{5.19E+07}\\
6&600&{\bf 6099035.253}&{\bf 14057515.8}&{3.34E+07}&{1.76E+08}\\
7&700&{\bf 12529985.88}&{\bf 19223817.05}&{2.17E+07}&{3.30E+08}\\
8&800&{\bf 2458899.443}&{\bf 3026434.255}&{4.93E+07}&{447915705.3}\\
9&900&{\bf 15425187.77}&{\bf 17203793.87}&{1.65E+08}&{5.25E+08}\\
10&1000&{\bf 2127.217017}&{\bf 118189.0265}&{7.50E+07}&{5.26E+08}\\
\hline %inserts single line
\end{tabular}
\label{table2} % is used to refer this table in the text
\end{table*}

\begin{figure}[h]
\begin{center}
\includegraphics[width=0.5\textwidth]{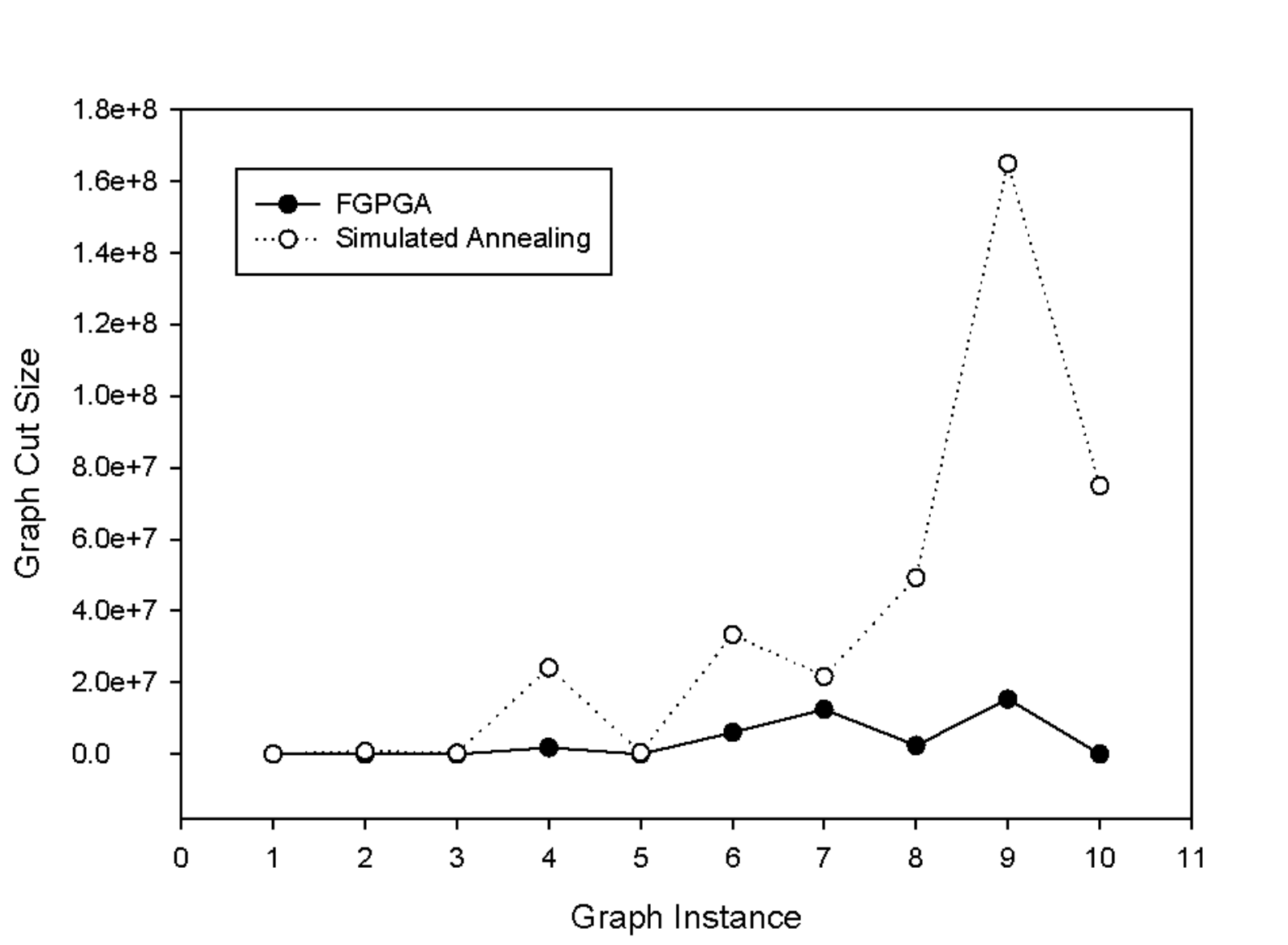}
\end{center}
\caption{Plot of best graph cut sizes of two algorithms FGPGA and simulated annealing for different graph instances, FGPGA produced a partition with cost zero for instance 1.\label{figAna1}}
\end{figure}

\begin{figure}[h]
\begin{center}
\includegraphics[width=0.5\textwidth]{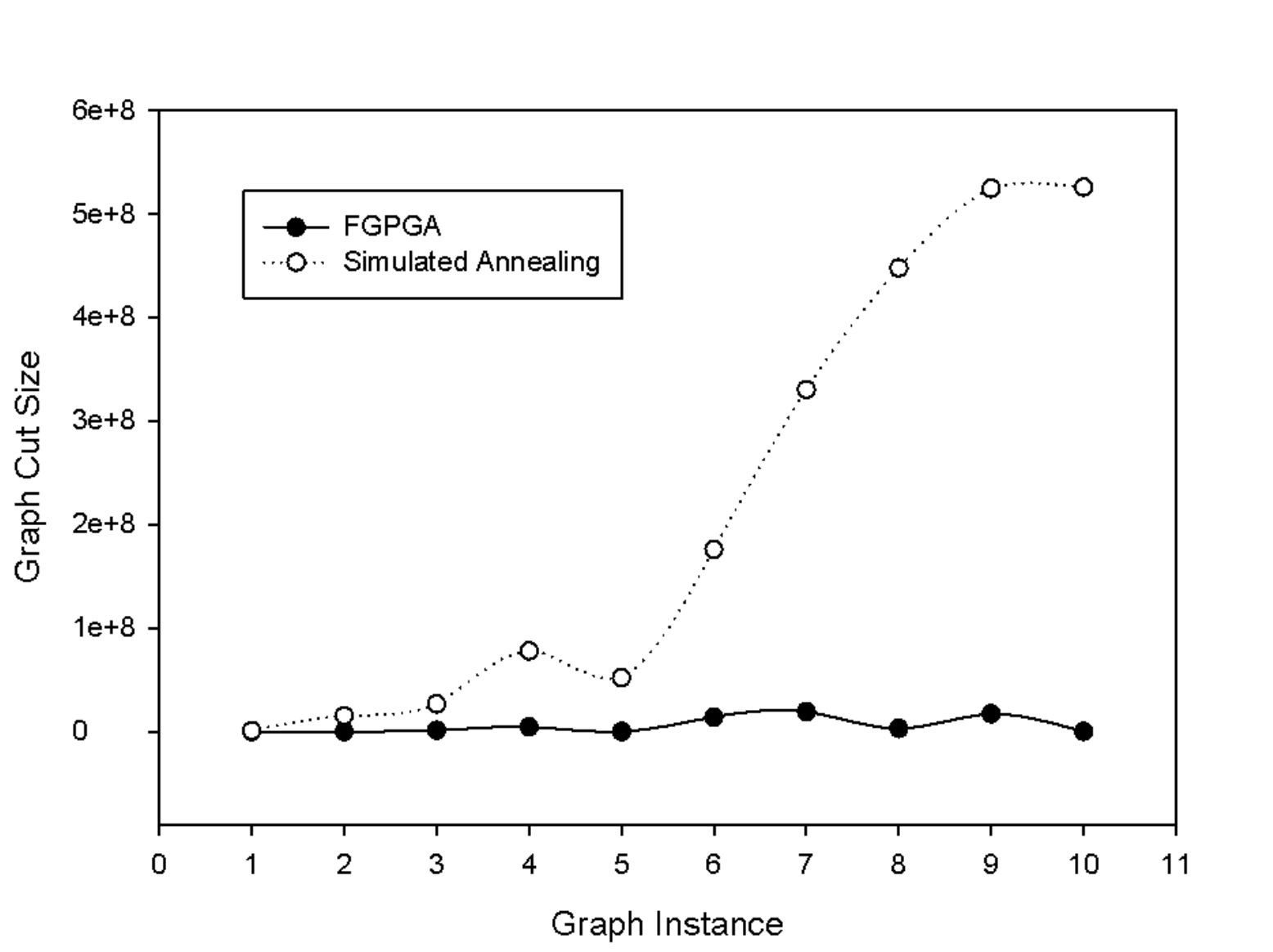}
\end{center}
\caption{Plot of average graph cut sizes of two algorithms FGPGA and simulated annealing for different graph instances, FGPGA produced a partition with cost zero for instance 1.\label{figAna2}}
\end{figure}

To show the significance of improvement of FGPGA over simulated annealing we plot the best and average of graph cut size of the best partitions found by both of the algorithms for each of the graph instances in Figure~\ref{figAna1} and Figure~\ref{figAna2} . We could see the difference in the quality of the partitions generated by both algorithms. Note that both algorithms produce partitions with relatively high cost for the graph instances with larger number of vertices. Both graph shows that the cut size increases with the increase in the number of vertices in an instance and the improvement of FGPGA also increases compared to the simulated annealing algorithm.

\subsection{Convergence}
\begin{figure}[h]
\begin{center}
\includegraphics[width=0.5\textwidth]{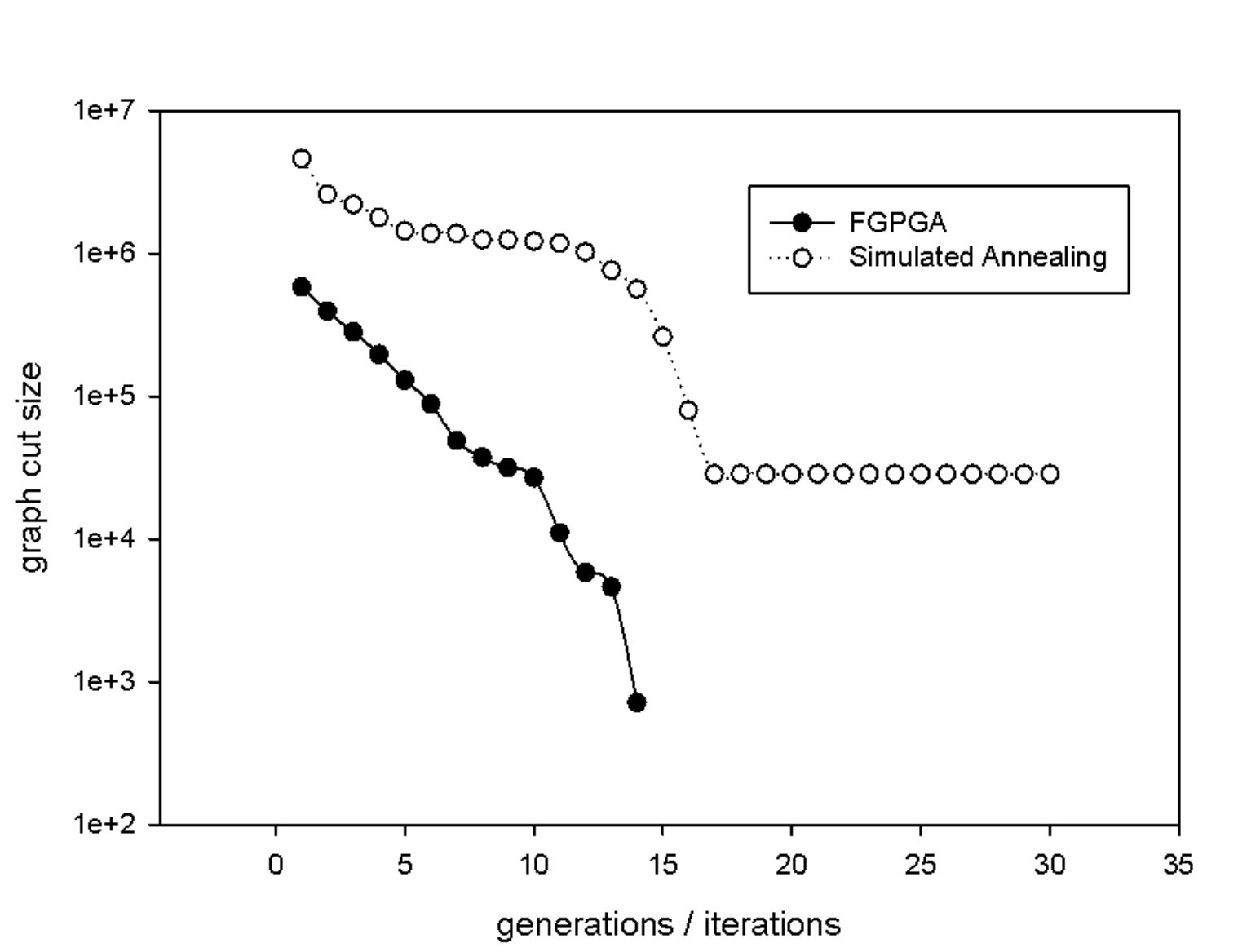}
\end{center}
\caption{Plot search progress for FGPGA and simulated annealing for graph size 100, FGPGA converges earlier than simulated annealing.\label{figProg}}
\end{figure}

Strength of our genetic approach is that it converges quickly compared to the simulated annealing algorithm. Search progress for two algorithms are shown in Figure~\ref{figProg}. We plot graph cut size of the best partition in each iteration of simulated annealing algorithms over the epochs and for each generation of genetic algorithms until they converge and terminate. 

The curves in Figure~\ref{figProg} indicates the earlier convergence of FGPGA. FGPGA converges quickly and terminates while simulated annealing fails to improve the quality of the partition even if the temperature is higher. This plot is showing the behavior of the algorithms for the graph instance 1 with number of vertices 100. Other graph instances show similar behavior.

\subsection{Effect of Greedy Mutation}
\begin{figure}[h]
\begin{center}
\includegraphics[width=0.5\textwidth]{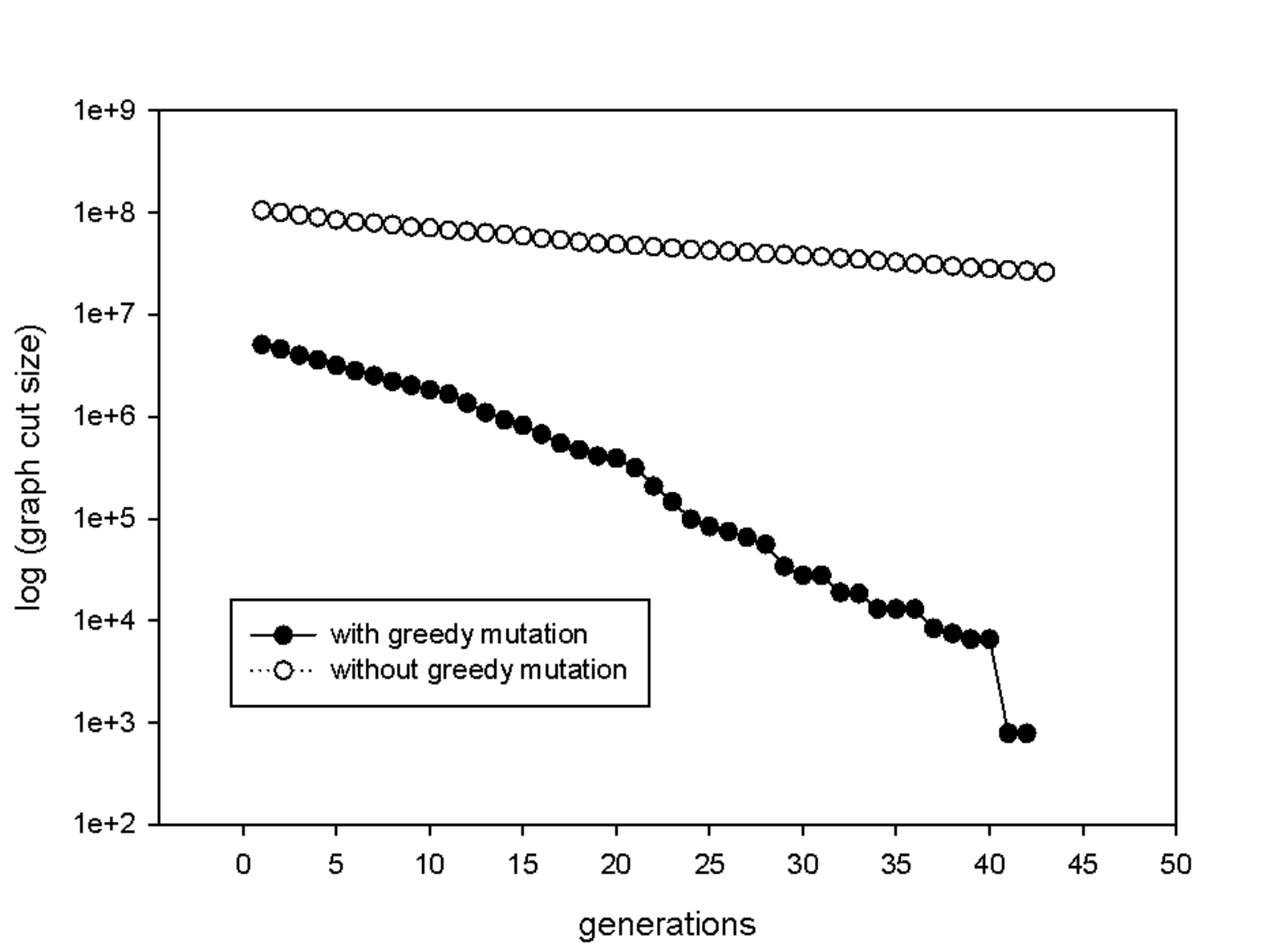}
\end{center}
\caption{Plot of search progress for FGPGA with greedy mutation and without greedy mutation for graph size 500, greedy mutation converges earlier.\label{figGM}}
\end{figure}
To show the effect of greedy mutation operator we prepare another version of the FGPGA algorithm by turning the greedy mutation off. In this version of FGPGA, the greedy mutation operator is replaced by a traditional random mutation operator. This operator selects the first valid mutation possible for any randomly selected gene value. The plot of logarithm of graph cut size is shown for each of the versions in Figure~\ref{figGM}. Figure~\ref{figGM} clearly shows the early convergence of greedy mutation operator. 

\section{Conclusion\label{secCon}} 
The recent emergence of mobile cloud computing leads to a new domain of challenges such as optimal deployment of software applications on the more resourceful infrastructure in the cloud instead of mobile hand set. Such problems can easily be modeled as a graph partitioning problem where a weighted graph of software components needed to be partitioned in a number of parts representing the available infrastructures (i.e., machines) in the cloud. Therefore, in this paper, we have introduced FGPGA, an efficient genetic approach for producing feasible graph partitions. Our method takes into account the heterogeneity and capacity constraints of the partitions that resemble real scenario of mobile cloud computing. Feasibility of the partitions generated is ensured by discarding over-sized partitions during the search. Experimental results exhibit the superiority of our proposed method on standard benchmark datasets over the state-of-the-art methods in terms of quality of partitions and feasibility of the solutions. Our algorithm is light weight and hence suitable for use in cloud architecture. In future we would like to deploy our algorithm to real architecture to investigate the performance of our proposed method for real world data.

\end{document}